\ifx\XeTeXversion\undefined
\pdfoutput=1
\fi
\documentclass[10pt,twocolumn,letterpaper]{article}
\ifdefined\cvprsubmission
  \usepackage[review]{cvpr}
  \def\paperID{*****} 
  \usepackage{xr-hyper}
\else
  \usepackage[pagenumbers]{cvpr}
\fi

\usepackage{graphicx}
\usepackage{amsmath}
\usepackage{amssymb}
\usepackage{booktabs}
\usepackage{multirow}
\usepackage{xcolor}

\definecolor{cvprblue}{rgb}{0.21,0.49,0.74}
\usepackage[pagebackref,breaklinks,colorlinks,allcolors=cvprblue]{hyperref}

\def\confName{CVPR}
\def\confYear{2026}

\newcommand{\ours}{Sparc4D}
\newcommand{\oursBase}{w/o stored texture}
\newcommand{\oursStatic}{w/o dynamic texture}
\newcommand{\best}[1]{\textbf{#1}}
\newcommand{\second}[1]{\underline{\textit{#1}}}

\newcommand{\blfootnote}[1]{%
  \begingroup
  \renewcommand{\thefootnote}{}\footnote{#1}%
  \addtocounter{footnote}{-1}%
  \endgroup
}

\ifdefined\cvprsubmission
\fi

\title{\ours{}: A Compact Explicit 4D Representation for Dynamic Scenes}
\author{Di Yang$^{1}$\qquad Zhihao Li$^{2}$ \qquad Yanhai Xiong$^{1,*}$\qquad Yufei Wang$^{2,*,\dagger}$\\[4pt]
$^{1}$College of William \& Mary\qquad $^{2}$SparcAI Inc.\\
{\tt\small dyang06@wm.edu\qquad zhihao.li@sparclab.ai\qquad yxiong05@wm.edu\qquad  yufei.wang@sparclab.ai}}
\begin{document}
\maketitle
\begin{abstract}
\ifdefined\cvprsubmission\else
\blfootnote{$^*$Corresponding author.\quad $^\dagger$Project lead.}%
\blfootnote{\textit{Preprint. Work in progress. 2026 Sep 26th.}}%
\fi
A compact dynamic-scene representation must retain both the surfaces seen over time and the appearance needed to render them from new viewpoints.
We present \ours{}, a feed-forward autoencoder that encodes a monocular video with known cameras into a sparse 4D scene state.
Static features are shared across the clip, while spatially anchored temporal slots store time-varying features with a median $5.0\times$ fewer values than per-frame storage and negligible change in the complete model's novel-view quality.
A sparse decoder produces 2D Gaussian surfels, while stored source pixels preserve fine texture through geometric re-projection.
The state includes one full source frame and dynamic-region pixels sampled every fourth frame, alongside learned features and sparse occupancy.
For a 32-frame MultiCamVideo clip, it averages 0.67M 32-bit-equivalent values on random windows and 0.66M on the first-32 protocol.
On first-32, \ours{} reaches 21.97\,dB, compared with 20.40\,dB for MoVieS.
On randomly placed windows, their PSNR scores are comparable.
Without fine-tuning on real data, \ours{} transfers to DyCheck and Neu3D, where stored texture improves LPIPS while slightly reducing PSNR.
Frozen \ours{} features also support a manipulation policy that reaches 81.3\% success on ManiSkill3 StackCube from 1,000 demonstrations.
\end{abstract}

\section{Introduction}
\label{sec:intro}

Feed-forward reconstruction methods can recover dynamic scenes from monocular video and render them from new viewpoints.
The resulting scene representation matters as much as the reconstruction procedure: it must retain information across time and remain small enough to store and reuse.
Pixel-aligned Gaussian methods such as MoVieS~\cite{lin2026movies} can produce large outputs when motion is materialized across a clip.
Compact Gaussian predictors such as C4G~\cite{kim2026c4g} reduce the number of primitives, highlighting the tradeoff between representation size and reconstruction fidelity.
This raises a representation question: how can a monocular video be encoded into a compact 4D scene state that supports explicit novel-view rendering?

Video tokenizers~\cite{yu2024magvit2,nvidia2025cosmos} compress image sequences, but their pixel-space outputs do not by themselves provide an explicit 3D rendering interface.
Sparse 3D autoencoders~\cite{xiang2024structured,xiang2025trellis2} provide such structure for individual assets.
Dynamic scenes require an additional choice: which information should be shared over time, and which should vary?
Storing a separate reconstruction for every frame repeats static content.
Compressing the entire scene into a coarse latent, on the other hand, makes fine appearance difficult to retain.
We address these two requirements separately.

\ours{} lifts a posed video into sparse 3D observations using estimated depth.
Static observations share one latent representation across the clip.
Dynamic observations are first encoded per frame, then compressed into a small number of temporal slots at each occupied latent cell.
A finer appearance code retains texture separately from the coarse latent.
The decoder reconstructs 2D Gaussian surfels~\cite{huang20242dgs} for explicit rendering.
Sparse occupancy is stored with the features, so it is included in the representation size.

Fine texture follows a second storage path: one full source frame preserves static detail, and dynamic-region pixels sampled every fourth frame refresh moving surfaces.
These stored pixels are re-projected onto decoded geometry, following image-based rendering and depth-image-based video coding~\cite{debevec1996modeling,buehler2001ulr,fehn2004dibr,tech2016overview}.
The complete state therefore combines learned spatial features with a sparse texture reservoir, whose cost is included in every main comparison.
Decoding reads only this state and the cameras, not the original video.

We evaluate reconstruction quality and state size on MultiCamVideo~\cite{bai2025recammaster}, including a C4G model fine-tuned on the same training split.
We also test reduced-output MoVieS variants and zero-shot transfer to DyCheck~\cite{gao2022dycheck} and Neu3D~\cite{li2022neural3dvideo}, analyze temporal compression, regional reconstruction, and keyframe texture, and train a manipulation policy on frozen encoder features to test whether the static and dynamic layers retain information useful for control.

The contributions are:
\begin{itemize}
    \item A sparse 4D scene representation that shares static features and compresses dynamic features into spatially anchored temporal slots.
    \item A surfel decoder that combines fine-grid appearance codes with stored static and dynamic texture, using explicit geometry to render both paths into new views.
    \item Experiments showing that temporal slots compress time-varying features with negligible change in novel-view reconstruction quality, and that \ours{} achieves competitive novel-view quality with a smaller state than MoVieS, 4DGT and C4G.
\end{itemize}

\section{Related Work}
\label{sec:related}

\paragraph{Dynamic scene reconstruction.}
Dynamic radiance fields model time-varying scenes through deformation~\cite{pumarola2021dnerf,park2021nerfies,park2021hypernerf}, scene flow~\cite{li2021nsff}, or image-based rendering~\cite{li2023dynibar}.
Gaussian splatting~\cite{kerbl20233dgs} supports deformable and time-varying primitives~\cite{wu20244dgs,yang2024deformable3dgs,luiten2024dynamic3dgs}.
Shape of Motion and MoSca regularize motion using low-dimensional bases or scaffolds~\cite{wang2024shapeofmotion,lei2025mosca}.
These methods optimize a scene-specific representation.
We instead learn an encoder and decoder shared across scenes.

\paragraph{Feed-forward reconstruction.}
Image-based reconstruction networks predict 3D representations in one pass~\cite{hong2024lrm,charatan2024pixelsplat,chen2024mvsplat,zhang2024gslrm,xu2025depthsplat}.
VGGT and MonST3R estimate geometry and cameras from static or dynamic inputs~\cite{wang2025vggt,zhang2025monst3r}.
For dynamic reconstruction, L4GM and STORM predict time-dependent Gaussians~\cite{ren2024l4gm,yang2025storm}.
MoVieS attaches motion to pixel-aligned Gaussians~\cite{lin2026movies}, 4DGT predicts space-time Gaussians from monocular video~\cite{xu2025fourdgt}, and C4G uses timestamp-conditioned queries to decode a compact Gaussian set~\cite{kim2026c4g}.
We compare with the latter three using their public weights and additionally fine-tune C4G on our training split at its original output budget.
Our focus is the encoded scene state and its size--quality tradeoff, rather than the number of output primitives alone.

\paragraph{Scene and video compression.}
Video tokenizers such as MAGVIT-v2 and Cosmos encode image sequences into spatiotemporal latents~\cite{yu2024magvit2,nvidia2025cosmos}.
TRELLIS, TRELLIS.2, Sparc3D and XCube use structured or hierarchical sparse representations for 3D assets~\cite{xiang2024structured,xiang2025trellis2,li2026sparc3d, ren2024xcube}.
SceneTok encodes static scenes into tokens decoded by a diffusion model~\cite{asim2026scenetok}.
We adapt the TRELLIS.2 shape autoencoder to dynamic scenes and add a temporal bottleneck based on learned queries~\cite{locatello2020slot,jaegle2021perceiver}.
The resulting state is decoded into surfels and rasterized, without a generative image-refinement or scene-completion module.
Structured 4D latents have also been used for robot planning~\cite{li2026structured4d}; Section~\ref{sec:control} relates a policy built on \ours{} features to their StackCube result.

\paragraph{Image-based rendering.}
View-dependent texture mapping, unstructured lumigraphs and learned image-based rendering use reference images to texture scene geometry~\cite{debevec1996modeling,buehler2001ulr,riegler2020fvs,wang2021ibrnet}.
Depth-image-based video coding similarly combines stored images and geometry to synthesize new views~\cite{fehn2004dibr,tech2016overview}.
Our texture reservoir follows this principle, but uses geometry reconstructed from the stored 4D state rather than transmitted depth maps.
For evaluation, we also adopt the distinction between observed and unobserved regions motivated by DyCheck~\cite{gao2022dycheck}.

\section{Method}
\label{sec:method}
\begin{figure*}[t]
  \centering
  \includegraphics[width=\linewidth]{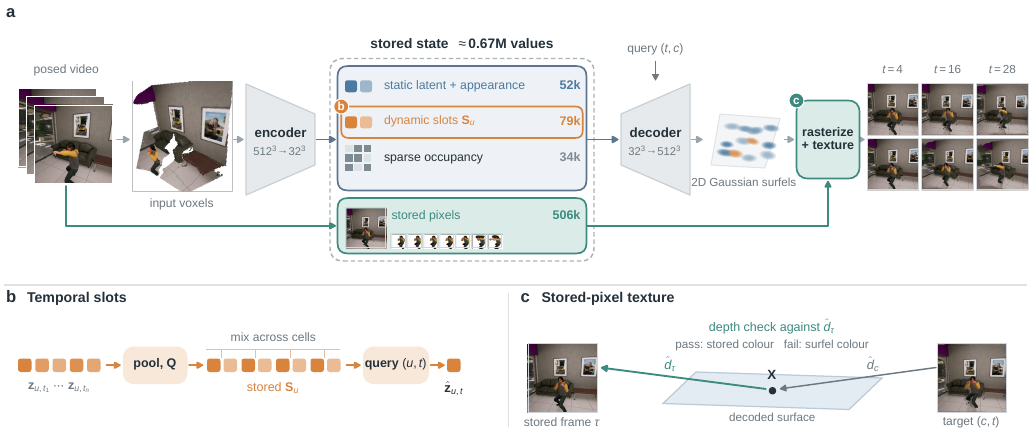}
  \caption{\textbf{Overview of \ours{}.} (a) Static observations are pooled over the clip and dynamic observations are kept per frame. A shared sparse encoder maps both to a learned state (blue box): static latents and appearance codes, dynamic temporal slots, and the sparse occupancy that guides decoder upsampling. Together with stored source pixels (green box), it forms the stored state (dashed box; mean values per clip, Table~\ref{tab:size}). At a query time $t$ and camera $c$, the decoder produces 2D Gaussian surfels that are rasterized and textured. Images are MultiCamVideo inputs and complete-model outputs; the voxels are the clip's input observations. The marked parts are detailed in (b) and (c). (b) Learned queries $\mathbf{Q}$ pool the per-frame features $\mathbf{z}_{u,t}$ of cell $u$; the slots are mixed across cells and stored as $\mathbf{S}_u$, and a query $(u,t)$ decodes $\hat{\mathbf{z}}_{u,t}$. Latent and appearance features use separate slots with 32 and 4 channels. (c) A target pixel is back-projected with the rendered depth $\hat d_c$ to $\mathbf{X}$ and takes the colour of a stored frame $\tau$ if the depth $\hat d_\tau$ rendered there agrees; otherwise it keeps the surfel colour. Geometry is unchanged.}

  \label{fig:overview}
\end{figure*}

Fig.~\ref{fig:overview}a summarizes \ours{}; panels b and c detail the temporal slots (Sec.~\ref{sec:slots}) and the stored texture (Sec.~\ref{sec:kf}).

\subsection{Scene observations}
\label{sec:input}

The input is a monocular clip of $T=32$ RGB frames $\{I_t\}$ at $256\times256$, with known intrinsics and world-to-camera poses.
We estimate depth using Depth Anything~3~\cite{lin2025da3}, conditioned on these cameras.
A pixel $\mathbf{x}$ is lifted into world coordinates as
\begin{equation}
\mathbf{p}_t(\mathbf{x})=
\mathbf{R}_t^\top\left(D_t(\mathbf{x})\mathbf{K}^{-1}\tilde{\mathbf{x}}-\mathbf{t}_t\right),
\label{eq:lift}
\end{equation}
where $D_t$ is estimated depth, $(\mathbf{R}_t,\mathbf{t}_t)$ is the camera pose, and $\tilde{\mathbf{x}}$ is the homogeneous pixel coordinate.
Each point carries RGB, a depth-derived normal and its position.

Static and dynamic observations serve different roles in the state.
We pool static points across the clip, but retain dynamic points separately for each frame.
To separate them, we test whether another frame sees through a point's expected position.
Foreground object masks~\cite{zheng2024birefnet} extend these motion seeds to whole objects, including intervals when an object briefly stops.
Appendix~\ref{app:preprocess} gives the test and mask parameters.

We voxelize the observations inside a scene-adaptive cube at $512^3$ resolution.
A larger, coarser cube stores static points outside this region.
An equirectangular environment map stores more distant observations by ray direction~\cite{debevec1998rendering}.
Pixels covered by neither surfels nor the map receive the source video's mean colour.
Each occupied voxel stores mean RGB, mean normal and a within-voxel point offset.
Octant colour residuals retain sub-voxel variation before encoding.

\subsection{Shared static features and temporal slots}
\label{sec:ae}

We initialize the sparse encoder--decoder from the TRELLIS.2 shape autoencoder~\cite{xiang2025trellis2}.
It maps a $512^3$ sparse voxel grid to a 32-channel latent on a $32^3$ grid.
Static, coarse-background and per-frame dynamic tensors use the same network weights.
The static and coarse latents are stored once per clip.
The decoder uses the input occupancy hierarchy for sparse upsampling; this hierarchy is part of the stored state.

\label{sec:slots}
The dynamic layer requires time-varying features, but need not store a separate latent for every frame.
Let $\mathcal{U}$ be the union of dynamic latent cells over the clip, and $\mathbf{z}_{u,t}$ the latent of cell $u$ at time $t$.
We compress the available per-frame features into two slots per cell:
\begin{equation}
\begin{aligned}
\mathbf{k}_{u,t}
 &= \mathbf{W}\mathbf{z}_{u,t}+\phi_t(t)+\phi_p(u),\\
\mathbf{A}_u
 &= \mathrm{Attn}\!\left(\mathbf{Q}+\phi_p(u),
                    \{\mathbf{k}_{u,t}\}_{t\in\mathcal{T}_u}\right),\\
\{\mathbf{S}_u\}_{u\in\mathcal{U}}
 &= \mathrm{Proj}\!\left(
    \mathrm{Mix}(\{\mathbf{A}_u\}_{u\in\mathcal{U}})
    \right),\\
\hat{\mathbf{z}}_{u,t}
 &= \mathrm{Dec}\!\left(
    \phi_t(t)+\phi_p(u),\mathbf{S}_u
    \right).
\end{aligned}
\label{eq:slots}
\end{equation}
Here $\mathcal{T}_u$ contains the observed times for cell $u$, and $\phi_t,\phi_p$ embed time and position.
Two learned queries $\mathbf{Q}$ first pool features within each cell.
Self-attention then mixes the slots across cells, and a projection reduces each slot to 32 channels.
To recover a latent, a query at $(u,t)$ attends to the two stored slots of $u$.
The decoded latent $\hat{\mathbf{z}}_{u,t}$ replaces $\mathbf{z}_{u,t}$ in the sparse decoder.
The state therefore stores $2|\mathcal{U}|$ latent vectors rather than all per-frame vectors.

\subsection{Appearance and surfel decoding}
\label{sec:app}

A cell of the $32^3$ latent grid spans $16^3$ input voxels.
We provide a finer path for texture by storing a 4-channel appearance code on the $128^3$ grid.
For encoder and decoder features $\mathbf{e}_u$ and $\mathbf{g}_u$ at this resolution,
\begin{equation}
\mathbf{a}_u=\mathbf{W}_e\,\mathrm{LN}(\mathbf{e}_u),
\qquad
\mathbf{g}_u\leftarrow\mathbf{g}_u+\gamma\mathbf{W}_d\mathbf{a}_u,
\label{eq:appearance}
\end{equation}
where $\mathrm{LN}$ is layer normalization and $\gamma$ is a fixed gain.
Static appearance codes are shared across the clip.
Dynamic appearance codes use the temporal-slot construction in Eq.~\ref{eq:slots}, with two 4-channel slots per union cell.

Each decoded voxel emits four 2D Gaussian surfels~\cite{huang20242dgs,pfister2000surfels} with degree-1 spherical-harmonics colour.
The head predicts centre offsets, scales, rotations and opacity as residuals from a camera-facing base configuration.
The base uses the source camera at the queried time; it does not use the target camera to redefine the surface.
We render the surfels with gsplat~\cite{ye2025gsplat}.
Appendix~\ref{app:head} gives the parameterization.
Dynamic decoding uses the occupancy observed at the queried time, so the model renders novel views at observed timestamps.

\subsection{Stored source-pixel texture}
\label{sec:kf}

The learned feature grids reconstruct surfaces and coarse appearance; a texture reservoir retains source detail without expanding every grid cell.
We store the first source frame in full and dynamic-region pixels every fourth frame.
The dynamic region is the decoded dynamic layer's source-view opacity above 0.5, dilated by two pixels.
Pixels already stored in the first frame are not counted again.
Both components are part of \ours{}, rather than optional test-time inputs.

For a target camera $c$ at time $t$, rasterization gives a colour image and median depth $\hat d_c$.
We back-project a target pixel using $\hat d_c$ and project the resulting point into a keyframe.
If its projected depth agrees with the depth rendered in that keyframe view, we use the sampled keyframe colour.
Otherwise, we retain the surfel colour.
Both depth maps come from the decoded state; no target-view depth estimate is needed for this operation.

Static pixels use the first source frame.
Dynamic pixels use the two stored dynamic frames that bracket the query time, each with the same projection and depth-consistency test, and blend the accepted colours by temporal distance.
This reuses colour at a reconstructed world position; it does not estimate motion or transport surfaces between timestamps.
Samples outside the stored region or failing the test retain the surfel colour.
Stored texture changes colour only, not rendered geometry or coverage.
The projection and consistency test are specified in Appendix~\ref{app:keyframes}.

\subsection{Training}
\label{sec:train}

For each training sample, we select a 32-frame window and one source camera.
Its frames form the input.
We supervise frame 0, a random time $t$ and the two stored dynamic frames bracketing $t$, rendering the source view and four of the remaining nine cameras at each.
The image loss combines $L_1$, SSIM and LPIPS and is computed after stored texture is applied; replaced pixels supervise only opacity.
Depth and normal losses supervise geometry, while a temporal loss matches inter-frame image differences.
The temporal slots also receive feature-reconstruction losses.
Occupancy and KL losses follow TRELLIS.2~\cite{xiang2025trellis2,kingma2014vae}.

\ours{} is initialized from the pretrained TRELLIS.2 model and trained for 30{,}000 steps on randomly placed 32-frame windows, with one clip per GPU on eight A800 GPUs.
Appendix~\ref{app:training} gives the full objective, optimization settings, and ablation protocol.

\subsection{State size}
\label{sec:size}

We count the stored features, sparse occupancy, environment map, background colour and texture reservoir.
The unit is a \emph{32-bit-equivalent value}: feature values count as one each, while occupancy entropy in bits is divided by 32.
Each stored 8-bit RGB channel is conservatively counted as one 32-bit value.
This is an estimated representation-size ledger, not a measured compressed bitstream.
Shared network weights and externally supplied cameras are excluded.

Let $\mathcal{C}$ and $\mathcal{U}$ denote static/coarse cells and dynamic union cells on the $32^3$ grid.
Their $128^3$ counterparts store appearance.
With occupancy cost $B_{\mathrm{occ}}$ bits, environment texel set $\mathcal{E}$ and additional dynamic-pixel set $\mathcal{P}_{\mathrm{dyn}}$, the reported size is
\begin{align}
N={}&32|\mathcal{C}|+64|\mathcal{U}|
     +4|\mathcal{C}^{128}|+8|\mathcal{U}^{128}| \nonumber\\
   &+B_{\mathrm{occ}}/32+4|\mathcal{E}|+3+3\cdot256^2
     +3|\mathcal{P}_{\mathrm{dyn}}|.
\label{eq:size}
\end{align}
Occupancy is estimated from octree symbols.
For dynamic occupancy, we use the smaller estimate from independent frames or consecutive-frame XOR masks.
Appendix~\ref{app:storage} details this accounting.
Table~\ref{tab:size} separates the learned state from its stored texture.

\section{Experiments}
\label{sec:exp}

We evaluate four questions: the reconstruction quality supported by a compact state, whether temporal slots reduce time-varying storage without sacrificing reconstruction quality, zero-shot transfer from synthetic training data to real videos, and whether the learned features support robotic control.

\subsection{Data and protocol}
\label{sec:protocol}

\paragraph{Dataset and split.}
MultiCamVideo (MCV)~\cite{bai2025recammaster} contains 3{,}400 Unreal Engine scenes.
Each scene is rendered at four focal lengths by ten synchronized cameras for 81 frames at 15\,fps.
A scene--focal-length pair forms one clip, giving 13{,}600 clips.
Frames are resized to $256^2$.

The held-out test set contains 101 scenes, selected by clustering DINOv2 scene features~\cite{oquab2024dinov2}, with all four focal lengths of each scene, giving 404 clips.
No test scene is used for training.
All MultiCamVideo results in this paper use this test set.
Appendix~\ref{app:split} gives the construction details.

\paragraph{Windows and cameras.}
The cameras coincide at the start of each video and then move apart.
We therefore evaluate two window placements.
The \emph{first-32} protocol uses frames 0--31.
The \emph{random-window} protocol fixes one randomly sampled start in $[0,49]$ for each clip.
The source camera is fixed by the clip's test-list index modulo ten; the other nine cameras are targets.
All methods render every target at all 32 timestamps, giving 116{,}352 target images per protocol.

\paragraph{Metrics.}
We report image-averaged PSNR, SSIM~\cite{wang2004ssim} and AlexNet-LPIPS~\cite{zhang2018lpips}.
Alongside full-image scores, we evaluate pixels classified as observed by a method-independent visibility mask.
The mask uses estimated depth and foreground segmentation; dynamic foreground requires same-time source visibility.
A third setting, \emph{uniform fill}, assigns the same source-mean colour to all methods outside the mask.
These settings separate reconstruction on observed regions from differences in unobserved-region rendering.
Appendix~\ref{app:metrics} specifies the masks and masked metrics.

\paragraph{Baselines and comparison scope.}
We evaluate MoVieS~\cite{lin2026movies}, 4DGT~\cite{xu2025fourdgt} and C4G~\cite{kim2026c4g} with released weights.
Because their original training data differ, these models are zero-shot references rather than a controlled same-training-data comparison.
To reduce this confound at a comparable representation scale, we additionally fine-tune C4G on the MultiCamVideo training split for 30{,}000 steps while retaining its 2{,}048-Gaussian output budget.
\ours{} is trained for the same number of steps on the same split.
The two methods retain their respective initializations, input protocols, and architectures.
All methods render at $256^2$; input settings and coordinate alignment are given in Appendix~\ref{app:baselines}.
Following the VideoTok4D convention~\cite{chen2026videotok4d}, \#Floats counts FP32 per-sequence state for 32 frames, excluding shared network weights and materializing time-conditioned Gaussians at every evaluated timestamp.

\subsection{Novel-view reconstruction}
\label{sec:main}
\begin{table*}[t]
\centering
\setlength{\tabcolsep}{3.2pt}
\resizebox{\textwidth}{!}{%
\begin{tabular}{lcr ccc c ccc ccc}
\toprule
& Trained & & \multicolumn{3}{c}{Full image} & & \multicolumn{3}{c}{Co-visible} & \multicolumn{3}{c}{Uniform fill} \\
\cmidrule(lr){4-6}\cmidrule(lr){8-10}\cmidrule(lr){11-13}
Method & on MCV & \#Floats & PSNR$\uparrow$ & SSIM$\uparrow$ & LPIPS$\downarrow$ & AbsRel$\downarrow$ & PSNR$\uparrow$ & SSIM$\uparrow$ & LPIPS$\downarrow$ & PSNR$\uparrow$ & SSIM$\uparrow$ & LPIPS$\downarrow$ \\
\midrule
MoVieS & -- & 719M & \second{17.26} & \second{0.512} & \best{0.366} & 0.615 (0.272) & \second{18.55} & \second{0.530} & \second{0.296} & \second{16.43} & \second{0.498} & \second{0.411} \\
4DGT & -- & 26.9M & 14.35 & 0.403 & 0.550 & 0.640 (0.370) & 15.14 & 0.411 & 0.538 & 14.49 & 0.412 & 0.546 \\
C4G & -- & \second{0.92M} & 13.64 & 0.369 & 0.596 & 0.543 (0.450) & 14.76 & 0.394 & 0.566 & 14.21 & 0.393 & 0.572 \\
C4G & \checkmark & \second{0.92M} & 16.51 & 0.466 & 0.584 & \second{0.356 (0.346)} & 17.14 & 0.470 & 0.561 & 15.65 & 0.444 & 0.564 \\
\midrule
\ours{} & \checkmark & \best{0.67M} & \best{17.39} & \best{0.519} & \second{0.370} & \best{0.259 (0.170)} & \best{19.02} & \best{0.541} & \best{0.290} & \best{16.62} & \best{0.509} & \best{0.405} \\
\bottomrule
\end{tabular}}
\caption{\textbf{Novel-view synthesis on MultiCamVideo} (404 random-window test clips, 9 target cameras $\times$ 32 frames, $256^2$). Best and second-best results are shown in bold and underlined italics, respectively. \ours{} denotes the complete method, including stored texture. \#Floats follows~\cite{chen2026videotok4d}: time-conditioned Gaussian outputs are materialized at every timestamp, and other methods count their per-sequence state (Eq.~\ref{eq:size} for ours); shared network weights are excluded. C4G is shown with released weights and after fine-tuning for 30{,}000 steps on the MCV training split, the number of steps used to train \ours{}. AbsRel uses DA3 pseudo ground truth, also our input depth; parentheses give median-scale-aligned error. Co-visible metrics use the shared visibility mask. Uniform fill assigns the same colour outside that mask for all methods.}
\label{tab:main}
\end{table*}

On random windows, \ours{} reaches 17.39\,dB with a 0.67M-value state (Table~\ref{tab:main}).
MoVieS obtains 17.26\,dB; the paired per-clip difference is $+0.12\pm0.18$\,dB (95\% CI).
On first-32, \ours{} reaches 21.97\,dB versus 20.40\,dB for MoVieS (Table~\ref{tab:rd}).

On co-visible pixels, \ours{} obtains 19.02\,dB and 0.290 LPIPS, compared with 18.55\,dB and 0.296 for MoVieS.
The paired LPIPS difference is $-0.005\pm0.008$, so these perceptual scores are comparable.
Uniform-fill evaluation preserves the PSNR ordering.
With a smaller state than MCV-fine-tuned C4G (0.67M versus 0.92M), \ours{} has higher full-image PSNR and lower LPIPS.
Section~\ref{sec:analysis} separates the contributions of learned features and stored texture.

\begin{figure*}[t]
  \centering
  \includegraphics[width=\linewidth]{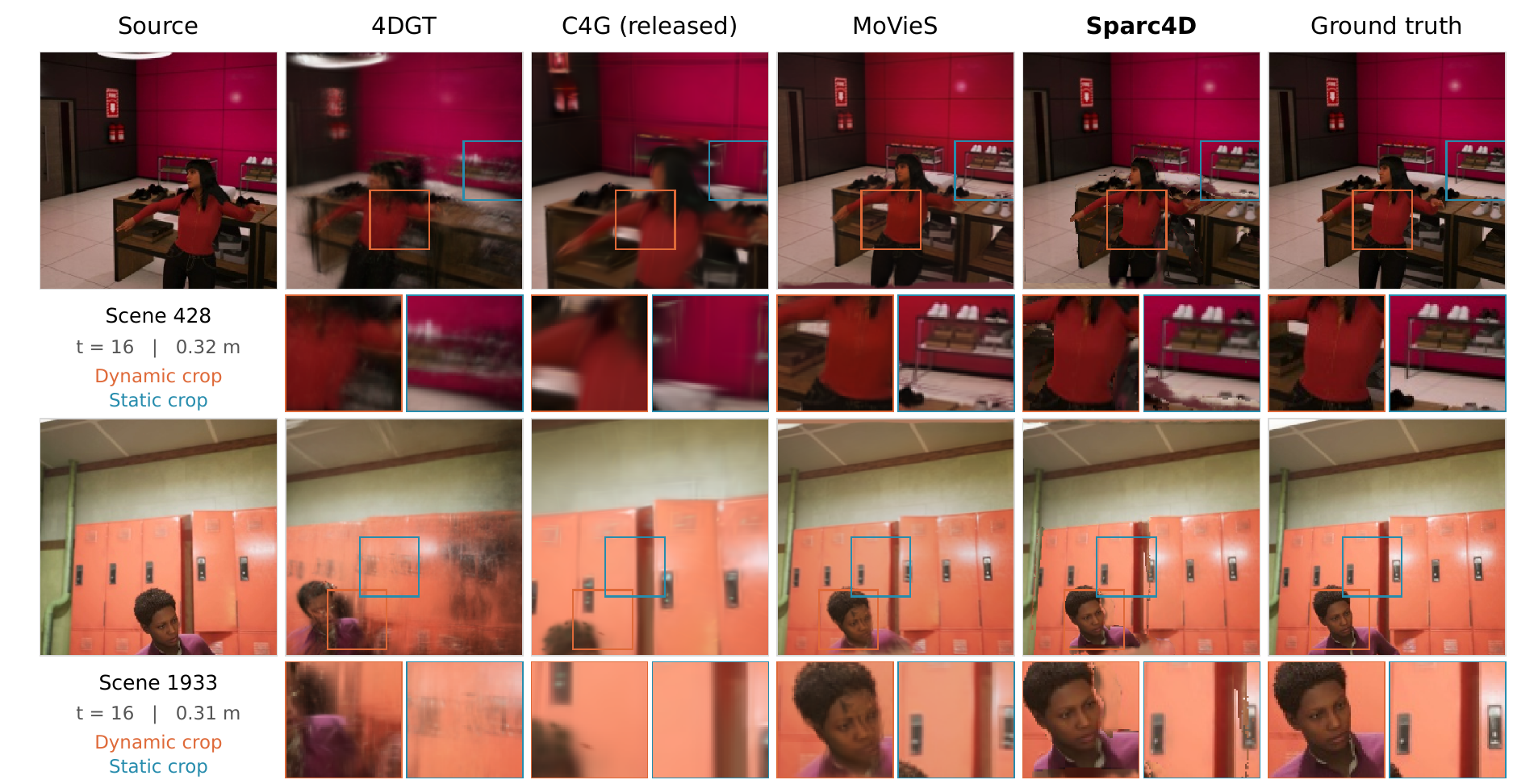}
  \caption{\textbf{Qualitative comparison, first-32 protocol.} Two fixed test scenes at frame 16, using the median-distance target camera. Orange and blue boxes mark dynamic and static $64^2$ crops, enlarged below at identical coordinates for all methods. \ours{} is the complete model; baseline images use released weights. Camera distance is shown in metres.}
  \label{fig:qual_first32}
\end{figure*}

\begin{figure*}[t]
  \centering
  \includegraphics[width=\linewidth]{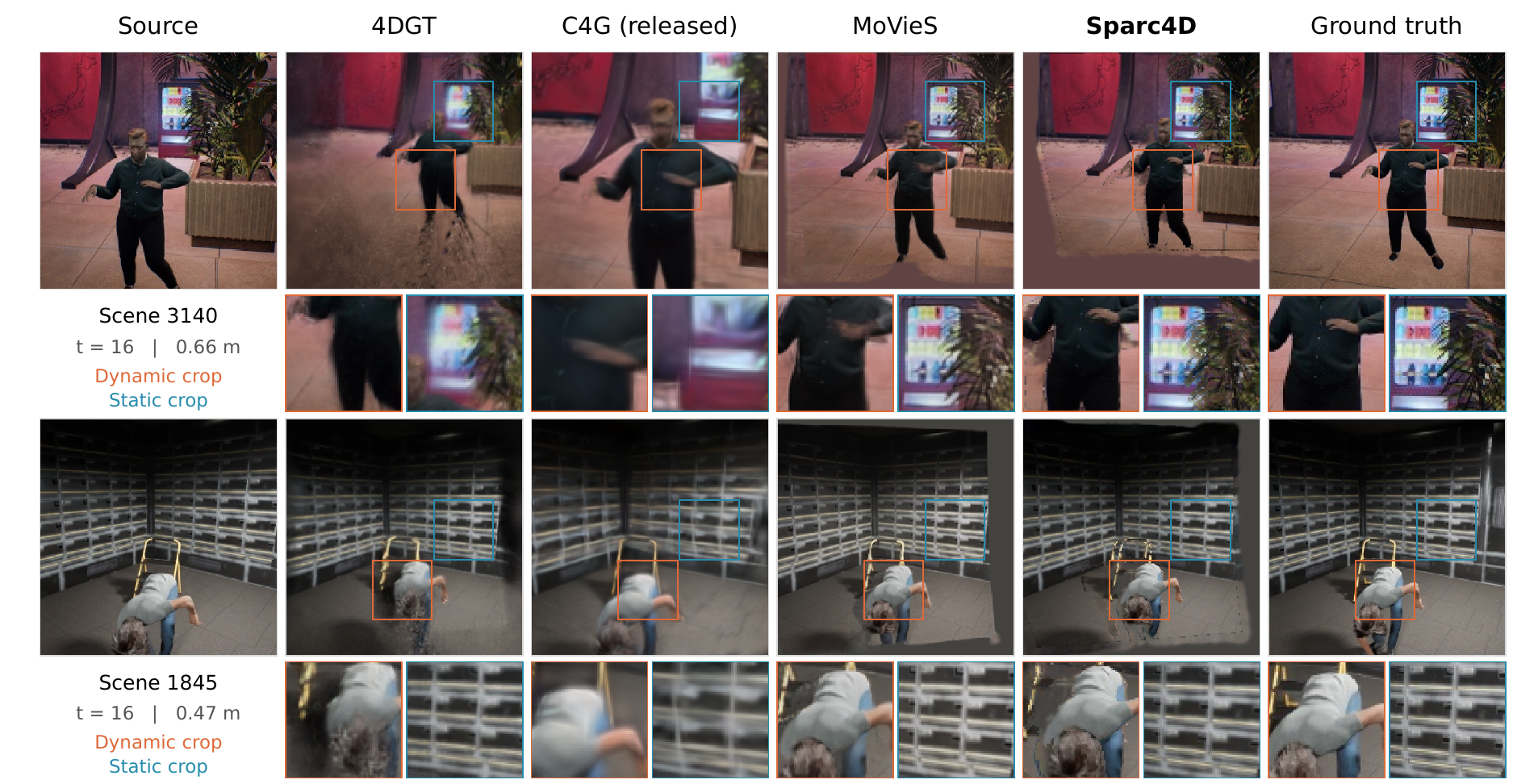}
  \caption{\textbf{Qualitative comparison, random-window protocol.} Same layout and method settings as Fig.~\ref{fig:qual_first32}.}
  \label{fig:qual_rand}
\end{figure*}

Figures~\ref{fig:qual_first32} and~\ref{fig:qual_rand} compare the complete method with the released baselines on fixed examples selected near the median per-clip PSNR difference.
The same scenes and crop locations are used for all methods; Appendix~\ref{app:qualitative} gives the selection procedure.

\subsection{Size--quality comparison}
\label{sec:rd}

We evaluate whether reducing the output of a pixel-aligned reconstructor can reach a similar size--quality tradeoff.
We give MoVieS fewer input frames or a lower input resolution, or prune its Gaussians by opacity, and evaluate all variants on the first-32 protocol (Table~\ref{tab:rd}).

With four input frames, MoVieS obtains 20.04\,dB with 28.5M output values; its single-frame variant obtains 18.97\,dB with 7.1M.
\ours{} reaches 21.97\,dB and 0.199 LPIPS with 0.66M values, about 11$\times$ smaller than the single-frame variant.
Lowering the input resolution or pruning by opacity reduces quality more strongly at comparable sizes.
The reduced-frame variants receive fewer input observations than our 32-frame model.
\begin{table}[tbp]
\centering
\small
\setlength{\tabcolsep}{4pt}
\caption{\textbf{Size and quality} on the first-32 protocol. MoVieS uses its native motion-parameter output count here, rather than the time-materialized count in Table~\ref{tab:main}; Appendix~\ref{app:storage} explains both. Our size includes all stored texture. MoVieS variants use $f$ inputs at resolution $r$, or retain the $k$ highest-opacity Gaussians.}
\label{tab:rd}
\begin{tabular}{lrccc}
\toprule
Method & Size & PSNR$\uparrow$ & SSIM$\uparrow$ & LPIPS$\downarrow$ \\
\midrule
MoVieS, 32f, 224  & 228M  & 20.40 & 0.643 & 0.207 \\
MoVieS, 8f, 224   & 57.0M & 20.28 & 0.642 & 0.203 \\
MoVieS, 4f, 224   & 28.5M & 20.04 & 0.628 & \second{0.201} \\
MoVieS, 2f, 224   & 14.3M & 19.61 & 0.619 & 0.219 \\
MoVieS, 1f, 224    & 7.1M  & 18.97 & 0.619 & 0.232 \\
MoVieS, 8f, 112   & 14.3M & 16.90 & 0.435 & 0.558 \\
MoVieS, top 200k        & 28.4M & 16.25 & 0.485 & 0.491 \\
MoVieS, top 50k         & 7.1M  & 14.96 & 0.434 & 0.642 \\
\midrule
\ours{}                 & \best{0.66M} & \best{21.97} & \best{0.686} & \best{0.199} \\
\bottomrule
\end{tabular}
\end{table}

\subsection{Transfer to real videos}
\label{sec:dycheck}

We evaluate zero-shot transfer to DyCheck iPhone~\cite{gao2022dycheck} and Neu3D~\cite{li2022neural3dvideo}(Table~\ref{tab:zeroshot}) without real-data fine-tuning.
DyCheck uses official co-visibility masks, whereas Neu3D uses full-image metrics; Appendix~\ref{app:dycheck} and Appendix~\ref{app:neu3d} give the full protocols.
\begin{table*}[t]
\centering
\setlength{\tabcolsep}{4pt}
\resizebox{\textwidth}{!}{%
\begin{tabular}{llr ccc ccc c}
\toprule
& & & \multicolumn{3}{c}{DyCheck} & \multicolumn{3}{c}{Neu3D (9 targets)} & Neu3D cam00 \\
\cmidrule(lr){4-6}\cmidrule(lr){7-9}\cmidrule(lr){10-10}
Method & Training data & \#Floats & mPSNR$\uparrow$ & mSSIM$\uparrow$ & mLPIPS$\downarrow$ & PSNR$\uparrow$ & SSIM$\uparrow$ & LPIPS$\downarrow$ & PSNR$\uparrow$ \\
\midrule
MoVieS & 8 public datasets ($\sim$174K scenes) & 719M & \best{13.95} & \second{0.345} & \best{0.488} & 15.61 & 0.445 & 0.304 & 17.01 \\
4DGT & $\sim$300\,h real monocular video & 26.9M & 11.84 & 0.277 & 0.599 & \best{17.33} & \best{0.530} & \best{0.281} & \best{19.51} \\
C4G & Spring, Kubric, RE10K & \second{0.92M} & 13.52 & \best{0.351} & 0.634 & 11.78 & 0.285 & 0.588 & 13.41 \\
C4G (MCV fine-tuned) & + MultiCamVideo & \second{0.92M} & 13.34 & 0.340 & 0.747 & 15.75 & 0.443 & 0.579 & 16.21 \\
\midrule
\ours{} & MultiCamVideo (synthetic) & \best{0.67M} & \second{13.82} & 0.323 & \second{0.543} & \second{16.57} & \second{0.465} & \second{0.291} & \second{18.54} \\
\bottomrule
\end{tabular}}
\caption{\textbf{Zero-shot generalization to real videos.}
Best and second-best results are shown in bold and underlined italics.
None of the methods is fine-tuned on DyCheck or Neu3D.
DyCheck uses official co-visibility evaluation; Neu3D reports nine target cameras and cam00 separately.
\#Floats repeats the MCV reference size for context rather than a measured real-data average.}
\label{tab:zeroshot}
\end{table*}

On DyCheck, \ours{} reaches 13.82\,dB mPSNR, close to MoVieS at
13.95\,dB, with higher mLPIPS (0.543 versus 0.488).
On Neu3D, it reaches 16.57\,dB across nine target cameras and
18.54\,dB on cam00; 4DGT reaches 17.33 and 19.51\,dB.

Table~\ref{tab:texture_transfer} isolates the effect of stored texture
using identical model weights and inputs.
On both real datasets, additional texture progressively improves LPIPS
while slightly reducing PSNR.

\subsection{Downstream manipulation}
\label{sec:control}

We evaluate the learned scene features on StackCube in ManiSkill3~\cite{tao2025maniskill3}, where a robot must grasp the red cube and stack it on the green cube (Fig.~\ref{fig:stackcube}).
After adapting the autoencoder to ManiSkill scenes, we freeze its encoder and train a flow-matching action policy on static features and the three most recent dynamic states.
Four calibrated RGB-D views provide the observations; robot masks and causal SAM~2 tracking~\cite{ravi2024sam2} separate the static and dynamic layers.
This control interface uses per-frame latents without temporal-slot compression or stored texture.

\begin{table}[tbp]
\centering
\small
\setlength{\tabcolsep}{4pt}
\caption{\textbf{StackCube success and scene size.}
Tokens are active cells per timestamp; $d$ is their feature dimension.
Protocols and counting details are in Appendix~\ref{app:control}.}
\label{tab:control}
\begin{tabular}{lrrrr}
\toprule
Method & Demos & Tokens & $d$ & Success \\
\midrule
Li et al.~\cite{li2026structured4d} & 1,000 & $\sim$5,700 & 8 & 84.0\% \\
\ours{} features & 1,000 & 661 & 32 & 81.3\% \\
\quad + additional demos & 3,000 & 661 & 32 & 84.0\% \\
\bottomrule
\end{tabular}
\end{table}

With 1,000 demonstrations, the policy reaches 81.3\% success, close to the 84.0\% reported by Li et al.~\cite{li2026structured4d}; adding 2,000 demonstrations raises it to 84.0\% (Table~\ref{tab:control}).
Our scene representation averages 426 static and 236 dynamic tokens, with the static features shared across the observation history.
The comparison uses each method's camera and control protocol.
The rollout in Fig.~\ref{fig:stackcube} illustrates control from these features.

\begin{figure}[tbp]
\centering
\setlength{\tabcolsep}{0pt}
\begin{tabular}{@{}cccc@{}}
\includegraphics[width=0.24\columnwidth,trim=344bp 56bp 278bp 40bp,clip]{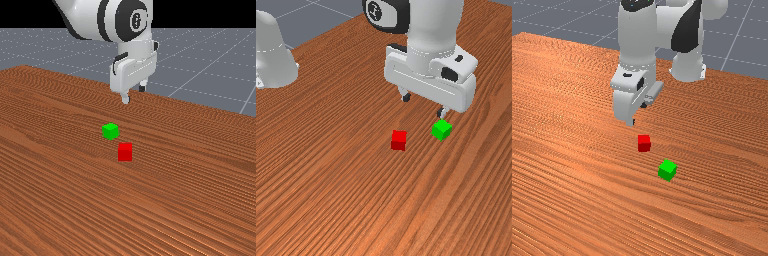} &
\includegraphics[width=0.24\columnwidth,trim=344bp 56bp 278bp 40bp,clip]{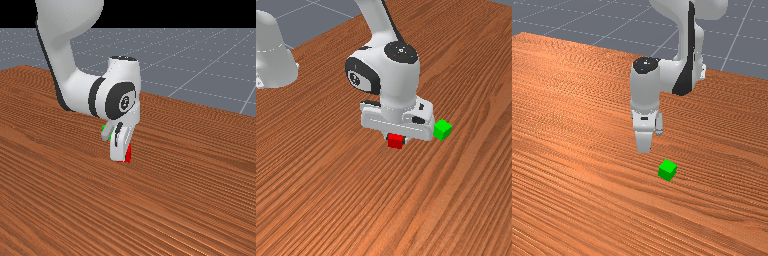} &
\includegraphics[width=0.24\columnwidth,trim=344bp 56bp 278bp 40bp,clip]{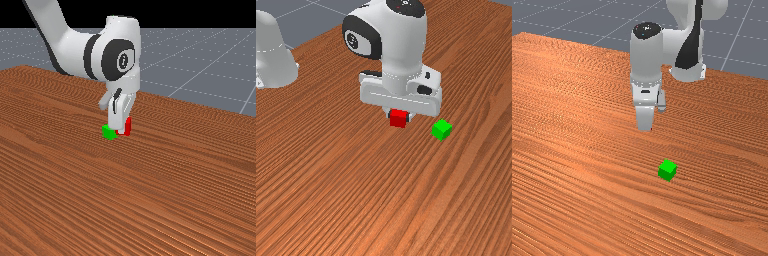} &
\includegraphics[width=0.24\columnwidth,trim=344bp 56bp 278bp 40bp,clip]{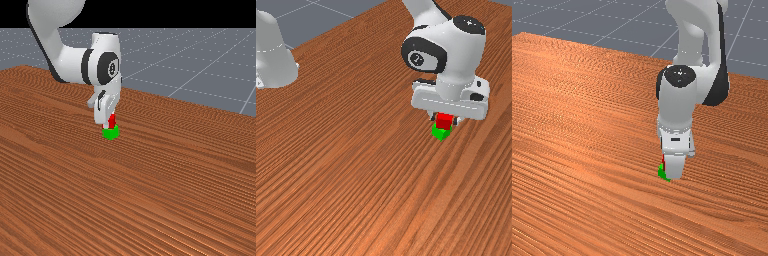} \\
\small\sffamily Approach & \small\sffamily Grasp & \small\sffamily Lift & \small\sffamily Stack
\end{tabular}
\caption{\textbf{StackCube manipulation.} A successful rollout of the policy trained on 1,000 demonstrations, shown from one fixed camera. The robot grasps the red cube and places it on the green cube.}
\label{fig:stackcube}
\end{figure}

\subsection{Representation analysis}
\label{sec:analysis}

\paragraph{Ablations.}
We first vary the number of temporal slots while holding the checkpoint, training schedule and absence of appearance codes fixed (Table~\ref{tab:abl_slots}).
Increasing $K$ from 2 to 10 raises PSNR by 0.04\,dB on average (paired 95\% CI: $\pm$0.01\,dB), while dynamic-slot storage increases from 21.1K to 105.6K values per clip.
The small quality gain in this 12k-step setting motivates the two-slot configuration used by the model.

Each configuration in Table~\ref{tab:abl_components} is trained for 10k steps from a shared 20k-step geometry-only checkpoint; Appendix~\ref{app:ablation_configs} gives the details.

The appearance pathway improves PSNR by 0.18\,dB over geometry only, while increasing the appearance-code capacity adds another 0.06\,dB.
The motion mask changes PSNR by $-0.04$\,dB, and the combined configuration differs by only $+0.01$\,dB from the appearance-code configuration.

\paragraph{Stored-texture ablation.}
Table~\ref{tab:abl_texture} varies only the stored texture of the trained model.
Without stored texture, PSNR is 16.82\,dB and LPIPS 0.469, compared with 17.39\,dB and 0.370 for \ours{}; most of this difference comes from the first frame.
Shorter intervals change PSNR by at most 0.03\,dB but enlarge the state by up to $2.5\times$, and blending the two bracketing frames outperforms both single-frame reuse and no reuse across time.

\begin{table}[tbp]
\centering
\small
\setlength{\tabcolsep}{3pt}
\caption{\textbf{Stored-texture ablation} (404 random-window clips). All rows use the same 30k-step backbone. Interval is the spacing between stored dynamic-texture frames; the first full source frame is retained except in ``w/o stored texture.''}
\label{tab:abl_texture}
\resizebox{\columnwidth}{!}{\begin{tabular}{lrccc}
\toprule
Configuration & Size & PSNR$\uparrow$ & LPIPS$\downarrow$ & Dyn.\ LPIPS$\downarrow$ \\
\midrule
\oursBase{} & 0.17M & 16.82 & 0.469 & 0.345 \\
\oursStatic{} & 0.36M & 17.31 & 0.384 & 0.292 \\
Dynamic texture interval: 1 & 1.69M & 17.42 & 0.367 & 0.258 \\
Dynamic texture interval: 2 & 1.01M & 17.41 & 0.368 & 0.260 \\
Dynamic texture interval: 8 & 0.50M & 17.34 & 0.371 & 0.266 \\
Interval 4, no reuse across time & 0.67M & 17.33 & 0.380 & 0.284 \\
Interval 4, preceding frame only & 0.67M & 17.35 & 0.370 & 0.263 \\
\midrule
\ours{} (interval: 4) & 0.67M & 17.39 & 0.370 & 0.262 \\
\bottomrule
\end{tabular}}
\end{table}

\paragraph{Temporal compression.}
Table~\ref{tab:temporal_compression} compares temporal slots with per-frame dynamic features.
The slots need a median $5.0\times$ fewer values for these features, while paired PSNR changes by only $+0.006\pm0.005$\,dB.
Appendix~\ref{tab:temporal_storage} gives the component-wise storage
breakdown and compression factors.

\begin{table}[tbp]
\centering
\small
\setlength{\tabcolsep}{3.0pt}
\caption{\textbf{Temporal compression ablation} on random windows.
Both variants use the same trained weights and stored texture, which is included in the total state.
``Per-frame'' bypasses temporal-slot compression for dynamic latent and
appearance features; dynamic occupancy is unchanged.
Sizes are mean 32-bit-equivalent values per 32-frame clip.}
\label{tab:temporal_compression}
\resizebox{\columnwidth}{!}{%
\begin{tabular}{lccccc}
\toprule
& \multicolumn{2}{c}{Storage $\downarrow$}
& \multicolumn{3}{c}{Reconstruction} \\
\cmidrule(lr){2-3}\cmidrule(lr){4-6}
Representation
& \shortstack{Time-varying\\features}
& \shortstack{Total\\state}
& PSNR$\uparrow$
& \shortstack{Dyn.\\PSNR$\uparrow$}
& \shortstack{Dyn.\\LPIPS$\downarrow$} \\
\midrule
Per-frame
& 354.5K & 0.96M
& 17.38 & 16.23 & 0.263 \\
Temporal slots
& \textbf{79.4K} & \textbf{0.67M}
& 17.39 & 16.24 & 0.262 \\
\bottomrule
\end{tabular}}
\end{table}

\paragraph{Voxelization and appearance.}

Table~\ref{tab:ladder} locates where dynamic detail is lost.
Voxelizing current-frame dynamic pixels at $512^3$ lowers source-view dynamic PSNR to 18.12\,dB, compared with 27.27\,dB for direct pixel rendering, and the learned state recovers 20.75\,dB without stored texture.
At target views, \ours{} reaches 16.36\,dB in dynamic regions, compared with 16.76\,dB for MoVieS, which transports primitives from multiple input frames to the query time (Sec.~\ref{sec:limitations}).

\paragraph{Static and dynamic reconstruction.}
On first-32, stored texture matters most in static regions, where \ours{} reaches 25.41\,dB, compared with 21.43\,dB without it and 21.79\,dB for MoVieS (Table~\ref{tab:region}).
In dynamic regions, \ours{} reaches 19.32\,dB, compared with 18.36\,dB when only the first frame is stored, and exceeds MoVieS by $0.34\pm0.32$\,dB in paired PSNR with similar LPIPS (0.153 versus 0.150).

\paragraph{Texture benefit over time.}
The benefit of stored texture decreases later in the window (Table~\ref{tab:time}).
The same trend remains when all source frames are stored, suggesting that increasing source--target separation and geometric re-projection error, rather than keyframe spacing alone, limit the benefit.
Dynamic refresh helps primarily at earlier timestamps.

Appendix~\ref{sec:geom} provides an additional analysis of decoded depth consistency, where \ours{} more closely preserves the DA3 depth structure than the evaluated baselines.
\section{Limitations}
\label{sec:limitations}

The decoder uses per-frame dynamic occupancy and renders at observed timestamps; it does not transport dynamic surfaces across time, so content hidden from the source camera at a query time is not recovered from other frames.
Stored dynamic texture improves fine detail but does not remove this support limitation.
Its cross-time re-projection assumes locally consistent geometry rather than explicit motion, so depth errors and fast motion can produce incorrect texture matches.
State size varies with occupied volume and dynamic image area, and our entropy-based ledger is not an implemented codec.

\section{Conclusion}
\label{sec:conclusion}

\ours{} encodes a monocular video into a sparse explicit 4D state with shared static features and spatially anchored temporal slots.
Together with fine-grid appearance codes and stored source texture, this yields a compact state whose novel-view quality is competitive with the evaluated feed-forward reconstructors.
The model also transfers from synthetic training to DyCheck and Neu3D without real-data fine-tuning, and its frozen features support a StackCube manipulation policy.


{
\small
\bibliographystyle{ieeenat_fullname}
\bibliography{main}
}

\ifdefined\cvprsubmission\else
\clearpage
\appendix
\section{Observation preprocessing and decoding}
\label{app:preprocess}

\paragraph{Depth and points.}
We run pose-conditioned Depth Anything~3 on overlapping 32-frame chunks of the full video.
Consecutive chunks are aligned in scale using the median depth ratio on their overlap.
Normals are computed from depth gradients, and sky pixels flagged by the depth model are excluded from the point cloud.

\paragraph{Motion seeds.}
Let $(\mathbf{x}',z')$ be the projection of $\mathbf{p}_t(\mathbf{x})$ into another frame $t'$.
The point is marked dynamic if, for some $|t-t'|\in\{1,2,4,8,16\}$,
\begin{equation}
\frac{D_{t'}(\mathbf{x}')-z'}{\max(z',0.05\,\mathrm{m})}>0.08.
\label{eq:seethrough}
\end{equation}
We set $D_{t'}=\infty$ on sky.
Connected components smaller than 48 pixels are removed, and the mask is dilated by four pixels.

\paragraph{Object masks.}
RMBG-2.0 (BiRefNet~\cite{zheng2024birefnet}) provides foreground components.
An 8-connected component $c$ is selected if it contains at least $\max(16,0.02|c|)$ motion-seed pixels and covers at most 35\% of the image.
Selections propagate forward and backward to components overlapping a selected region by at least 30\% of their area.
Selected regions are dilated by two pixels.

\paragraph{Spatial layout.}
The scene cube has side length $s=\mathrm{clip}(3\tilde d,4\,\mathrm{m},32\,\mathrm{m})$, where $\tilde d$ is median camera-to-point distance.
Its centre is the median of the nearest 30\% of points.
A $512^3$ grid gives voxel widths of approximately 0.8--6.3\,cm.
A concentric 64\,m cube at the same grid resolution stores outlying static points with 12.5\,cm voxels.
Points beyond that cube contribute to a $1024\times512$ equirectangular map by viewing direction relative to the camera-trajectory centre.
Only sampled texels are stored.

\paragraph{Voxel features.}
Each voxel has nine base features: mean RGB, mean normal and mean within-voxel point offset.
For each of eight octants we append the difference between its mean RGB and the voxel mean, plus an occupancy flag.

\subsection{Surfel parameterization}
\label{app:head}
Each output voxel $\mathbf{v}$ (integer coordinates, voxel size $h$, cube origin $\mathbf{o}$) emits $K{=}4$ 2D Gaussian surfels~\cite{huang20242dgs,pfister2000surfels} with degree-1 spherical-harmonics colour. The head predicts residuals $(\boldsymbol{\delta},\boldsymbol{\ell},\hat{\mathbf{q}},a)$ around a base configuration that depends on the source camera centre $\mathbf{c}_t$ at the rendered time $t$:
\begin{align}
\boldsymbol{\mu} &= \mathbf{o} + h\,\big(\mathbf{v} + \tfrac12 + \tanh\boldsymbol{\delta}\big), \qquad
\alpha = \sigma(a + 2), \nonumber\\
\mathbf{s} &= \mathrm{cap}\Big(\rho\,\tfrac{\mathrm{softplus}(\boldsymbol{\ell}+b)}{\mathrm{softplus}(b)}\Big), \qquad \rho = \frac{0.55\,\|\mathbf{c}_t-\bar{\mathbf{v}}\|}{f}, \\
\mathbf{q} &= \mathrm{norm}\big(\hat{\mathbf{q}} + \mathbf{1}_w\big)\otimes \mathbf{q}_{\mathrm{face}}\big(\mathbf{c}_t-\bar{\mathbf{v}}\big), \nonumber
\end{align}
where $\bar{\mathbf{v}}$ is the voxel centre, $\rho$ is the world-space footprint of 0.55 pixels at focal length $f$, $b=\log(e-1)$, $\mathrm{cap}(s)=s\,s_{\max}/(s+s_{\max})$ with $s_{\max}$ four voxels, $\mathbf{1}_w$ the identity quaternion, and $\mathbf{q}_{\mathrm{face}}(\mathbf{n})$ the rotation that turns the surfel normal towards $\mathbf{n}$. With a zero-initialized output layer the head therefore renders each voxel as a camera-facing disk of about one pixel, providing the initial configuration for residual prediction. Coarse-cube voxels use the same head with their own voxel size and a minimum footprint of half a coarse voxel.

\subsection{Appearance and slot modules}
\label{app:modules}

$\mathbf{W}_d$ and the octant-feature input projection are zero-initialized and scaled by a fixed gain $\gamma=10$, so both residual pathways start at zero; $\mathbf{W}_e$ uses the default initialization.
Gradients in the gained layers are scaled by $1/\gamma^2$ before clipping.

The temporal modules use width 256 and eight attention heads.
Two learned queries pool the observed tokens of each cell.
Three self-attention layers mix all slots, followed by projection to the stored channel dimension.
Three cross-attention layers decode each $(u,t)$ query.
Latent slots have 32 channels; appearance slots have 4.
Both receive the normalized feature-matching losses in Appendix~\ref{app:training}.

\subsection{Stored-texture re-projection}
\label{app:keyframes}

Additional-keyframe ablations store $k$ full frames at $\mathcal{K}=\{\mathrm{round}(31i/(k-1))\}_{i=0}^{k-1}$.

Given target depth $\hat d_c$, a pixel is back-projected to
$\mathbf{X}=\pi_c^{-1}(\mathbf{x},\hat d_c(\mathbf{x}))$ and projected into keyframe $\tau$ as $(\mathbf{y},z)=\pi_\tau(\mathbf{X})$, where $\pi$ denotes calibrated camera projection.
We accept the sampled colour $I_\tau(\mathbf{y})$ when
\begin{equation}
|z-\hat d_\tau(\mathbf{y})|<0.03z+0.02\,\mathrm{m},
\label{eq:kf}
\end{equation}
where $\hat d_\tau$ is rendered from the same decoded state in the keyframe view.
Otherwise the surfel colour is retained.
Static pixels try keyframes in order of source-camera distance to the target camera.
From full keyframes, pixels where the dynamic layer alone has opacity above 0.5 take colour only at the keyframe's own timestamp; at other times they use the dynamic texture below.

\paragraph{Dynamic texture reservoir.}
Dynamic texture is stored at $t=4,8,\ldots,28$.
A dynamic pixel at time $t$ blends the stored frames $\tau^-\le t<\tau^+$ by temporal proximity, using only samples that satisfy Eq.~\ref{eq:kf} and have target opacity above 0.5; source depth is rendered at each stored timestamp.
Frame 0 serves as $\tau^-$ for $t<4$, and only $\tau^-$ is available for $t>28$.
Texture is applied before environment-map compositing.

\section{Training details}
\label{app:training}

The supervised timestamps are $\{0,t,\tau^-,\tau^+\}$, padded with random timestamps when they coincide or $\tau^+$ falls outside the window, and the four target cameras are drawn at random from the nine non-source cameras.
The loss is
\begin{align}
\mathcal{L} &= \tfrac{1}{4}\textstyle\sum_{q}\mathcal{L}_q
+ 0.5\,\mathcal{L}_{\Delta t}
+ \mathcal{L}_{\text{lat}}
+ \mathcal{L}_{\text{lat}}^{\text{app}}
\nonumber\\
&\quad
+ 0.1\,\mathcal{L}_{\text{occ}}
+ 10^{-6}\,\mathcal{L}_{\text{KL}}, \\
\mathcal{L}_q
&=
\mathcal{L}_{\text{rgb}}^{\text{src}}
+ \tfrac{1}{4}\textstyle\sum_{v}\mathcal{L}_{\text{rgb}}^{v}
\nonumber\\
&\quad
+ 0.3\,\big(
\mathcal{L}_{d}^{\text{src}}
+ \tfrac{1}{4}\textstyle\sum_{v}\mathcal{L}_{d}^{v}
\big)
+ 0.05\,\mathcal{L}_{n},
\end{align}
where $v$ runs over the four supervising target cameras.
We use
\[
\mathcal{L}_{\text{rgb}}
=
\|\hat I-I\|_1
+0.2(1-\mathrm{SSIM})
+\lambda_{\mathrm{LPIPS}}\,
\mathrm{LPIPS}_{\text{VGG}}
\]
on images composited over the mean-colour background after stored-texture replacement.

$\mathcal{L}_d$ is an $L_1$ loss on log depth against the per-camera depth estimate, restricted to pixels inside the scene cube and covered by surfels.
$\mathcal{L}_n$ is the 2DGS normal-consistency term~\cite{huang20242dgs}.
$\mathcal{L}_{\Delta t}
=
\mathrm{mean}_{a<b}
\|(\hat I_a-\hat I_b)-(I_a-I_b)\|_1$
penalizes temporal differences that do not match the ground truth.
$\mathcal{L}_{\text{lat}}$ and
$\mathcal{L}_{\text{lat}}^{\text{app}}$
are the relative squared errors between slot-decoded and per-frame latents and appearance codes; for the appearance code, the loss is evaluated on a random subset of $65{,}536$ rows.
$\mathcal{L}_{\text{occ}}$ and
$\mathcal{L}_{\text{KL}}$
are the occupancy and KL terms of TRELLIS.2~\cite{xiang2025trellis2,kingma2014vae}.

\paragraph{Final model.}
The final model combines object-level motion masks, 4-channel appearance codes with slot-compressed dynamic codes, and training with stored texture.

\paragraph{Optimization.}
We use AdamW~\cite{loshchilov2019adamw}
($\beta=(0.9,0.95)$, no weight decay),
a learning rate of $3\times10^{-5}$ with 200 warm-up steps and cosine decay to $10^{-5}$,
a $10\times$ learning rate for the slot modules, $\lambda_{\mathrm{LPIPS}}=0.6$, and global gradient-norm clipping at 1.0.
The model has 836M trainable parameters, almost all in the TRELLIS.2 backbone.

\subsection{Ablation configurations}
\label{app:ablation_configs}
\begin{table}[tbp]
\centering
\small
\setlength{\tabcolsep}{4pt}
\caption{\textbf{Number of temporal slots $K$} on MultiCamVideo (404 test clips, full image). All variants start from the same checkpoint and are trained for 12k steps without appearance codes; only $K$ differs. Dynamic-slot size is the mean number of 32-bit values per clip. $\Delta$PSNR is the per-clip paired difference to $K{=}2$ (mean $\pm$ 95\% CI).}
\label{tab:abl_slots}
\begin{tabular}{lrcccc}
\toprule
$K$ & \shortstack{Dyn.\ slot\\\#Floats} & PSNR$\uparrow$ & SSIM$\uparrow$ & LPIPS$\downarrow$ & $\Delta$PSNR \\
\midrule
2  & \best{21.1K}   & 16.76 & \second{0.492} & \second{0.475} & -- \\
3  & \second{31.7K} & \second{16.78} & \best{0.493} & \second{0.475} & $+$0.02 $\pm$ 0.01 \\
10 & 105.6K & \best{16.80} & \best{0.493} & \best{0.474} & $+$0.04 $\pm$ 0.01 \\
\bottomrule
\end{tabular}
\end{table}

\begin{table}[!htbp]
\centering
\small
\setlength{\tabcolsep}{1.5pt}
\caption{\textbf{Learned-architecture ablation}
(404 random-window clips).
Each row is trained for 10k steps from a shared 20k-step geometry-only checkpoint (30k steps in total).
$\Delta$PSNR gives paired per-clip differences to the corresponding parent configuration (mean $\pm$ 95\% CI).
All rows omit stored texture.}
\label{tab:abl_components}
\resizebox{\columnwidth}{!}{%
\begin{tabular}{lrcccc}
\toprule
Configuration & \#Floats & PSNR$\uparrow$ & SSIM$\uparrow$ & LPIPS$\downarrow$ & $\Delta$PSNR \\
\midrule
Geometry only & 76.2K & 16.76 & 0.492 & 0.475 & -- \\
\quad + appearance code & 0.87M & 16.94 & 0.500 & 0.448 & $+0.18\pm0.04$ \\
\quad + larger code & 3.52M & \best{16.99} & \best{0.502} & \second{0.442} & $+0.06\pm0.01$ \\
\quad + motion mask & 3.58M & \second{16.96} & \best{0.502} & 0.444 & $-0.04\pm0.01$ \\
\midrule
Combined config. & 0.42M & 16.95 & \second{0.501} & \best{0.430} & $+0.01\pm0.03$ \\
\bottomrule
\end{tabular}}
\end{table}

The appearance-code configuration adds a 16-channel $128^3$ code and sub-voxel colour input.
The larger-code configuration increases this to 32 channels and adds a 16-channel $256^3$ code, while the motion-mask configuration additionally uses object-level dynamic masks.
The combined configuration instead retains the 16-channel $128^3$ code and combines object-level masks, slot-compressed dynamic appearance, random windows across the 81-frame clips, and $\lambda_{\mathrm{LPIPS}}=0.6$.
The final model further reduces the appearance code to four channels and is trained with stored texture.

\section{Representation-size accounting}
\label{app:storage}
\begin{table}[tbp]
\centering
\small
\setlength{\tabcolsep}{2.5pt}
\caption{\textbf{State size of \ours{}} in thousands of 32-bit-equivalent values per 32-frame clip (random-window protocol; first-32 values in parentheses). Occupancy uses empirical entropy divided by 32. The first full frame adds $256^2\times3$ values; dynamic texture cost depends on the retained pixel area.}
\label{tab:size}
\begin{tabular}{lrr}
\toprule
Component & Mean & Median \\
\midrule
Static + coarse latent & 14.7 & 12.1 \\
Dynamic temporal slots & 22.7 & 16.3 \\
Static appearance code & 27.1 & 22.8 \\
Dynamic appearance slots & 56.7 & 38.5 \\
Occupancy, static + coarse & 6.3 & 5.1 \\
Dynamic occupancy (32f) & 27.8 & 19.9 \\
Environment map & 10.4 & 0.0 \\
\midrule
State without stored texture & 165.7 \,(158.8) & 129.7 \,(130.7) \\
Subtotal with first-frame texture & 362.3 \,(355.4) & 326.3 \,(327.3) \\
\midrule
Dyn.\ source pixels, every 4th fr. & 308.8 \,(302.0) & 256.0 \,(238.1) \\
\ours{} (complete state) & 671.2 \,(657.4) & 588.0 \,(568.9) \\
\bottomrule
\end{tabular}
\end{table}

Occupancy is represented as a $32^3$ bitmap followed by 8-bit child masks at four finer octree levels, and its cost is the zeroth-order empirical entropy of these symbols.
Each environment texel counts as four values.
Dynamic-texture masks are not stored, because the decoder recomputes them from the dynamic geometry; scene-layout metadata such as cube origin and scale are omitted.

\paragraph{Temporal-compression ablation.}
For Table~\ref{tab:temporal_compression}, the per-frame variant bypasses the slots at inference and stores dynamic latent and appearance features at every timestamp, at a cost of
\begin{equation}
N_{\mathrm{pf}}
=
32\sum_{t=1}^{T} |\mathcal{U}_t|
+
4\sum_{t=1}^{T} |\mathcal{U}^{128}_t|,
\label{eq:perframe-storage}
\end{equation}
where $\mathcal{U}_t$ and $\mathcal{U}^{128}_t$ are the occupied dynamic cells at time $t$ on the $32^3$ latent and $128^3$ appearance grids, respectively.
With temporal slots, the corresponding cost is
\begin{equation}
N_{\mathrm{slot}}
=
64|\mathcal{U}|
+
8|\mathcal{U}^{128}|,
\label{eq:slot-storage}
\end{equation}
where $\mathcal{U}$ and $\mathcal{U}^{128}$ are the temporal unions of occupied dynamic cells.

\begin{table}[!htbp]
\centering
\small
\setlength{\tabcolsep}{2.5pt}
\caption{\textbf{Temporal feature storage} over the 404 random-window test clips. Sizes are mean 32-bit-equivalent values per clip. Compression factors are computed per clip and reported as their median and mean. Total state includes stored texture.}
\label{tab:temporal_storage}
\resizebox{\columnwidth}{!}{%
\begin{tabular}{lrrrr}
\toprule
Component & Per-frame & Slots & Median & Mean \\
\midrule
Dynamic latent & 182.1K & 22.7K & $8.8\times$ & $8.7\times$ \\
Dynamic appearance & 172.5K & 56.7K & $3.5\times$ & $3.8\times$ \\
\midrule
Time-varying features & 354.5K & 79.4K & $5.0\times$ & $5.2\times$ \\
Total state & 0.96M & 0.67M & $1.4\times$ & $1.4\times$ \\
\bottomrule
\end{tabular}%
}
\end{table}

For the combined time-varying feature state, the 10th--90th percentile range of the per-clip compression ratio is $3.3\times$--$7.4\times$.

For MoVieS, the materialized output contains $1{,}605{,}632$ Gaussians at $32\times224^2$ input pixels.
The reduced-output study in Table~\ref{tab:rd} counts its native prediction tensors: 14 static parameters and $4\times32$ query-time motion parameters per Gaussian, or 142 values.
For the cross-method tables, we follow the VideoTok4D reporting convention~\cite{chen2026videotok4d}, which materializes the time-conditioned representation at every evaluated timestamp and gives 719M FP32 values for a 32-frame sequence.
For 4DGT we count the predicted Gaussian parameters.
For C4G we count the outputs of 2{,}048 Gaussians at each of the 32 queried timestamps.

\paragraph{Two size conventions.}
Table~\ref{tab:size_conv} adds the stored size, which counts what must be kept to render any observed timestamp.
\ours{} is smallest under this convention, whereas its dense decoded output, about 0.39M surfels per timestamp, exceeds that of C4G when materialized.

\begin{table}[!htbp]
\centering
\small
\setlength{\tabcolsep}{4pt}
\caption{\textbf{Representation size under two conventions}, in values per 32-frame clip. Materialized sizes use a common 14 values per primitive (position, scale, rotation, opacity and one RGB colour) at each of the 32 timestamps, regardless of each method's native colour model; with its degree-1 spherical-harmonics colour, \ours{} would count 22 values per surfel (277M); for \ours{} they are averaged over the 24 clips of Table~\ref{tab:runtime}. $^\dagger$Upper bound without lifetime culling. $^\ddagger$Estimated from the architecture: 12 context frames $\times$ 256 tokens $\times$ 2{,}048 channels.}
\label{tab:size_conv}
\begin{tabular}{lrr}
\toprule
Method & Stored & Materialized \\
\midrule
MoVieS & 228M & 719M \\
4DGT & 26.9M & 301M$^\dagger$ \\
C4G & 6.3M$^\ddagger$ & \best{0.92M} \\
\ours{} & \best{0.67M} & 176M \\
\bottomrule
\end{tabular}
\end{table}

\section{Runtime and memory}
\label{app:runtime}

Table~\ref{tab:runtime} reports runtime on one A800 GPU with batch size 1, as the median over 24 test clips.
Each clip has 32 input frames and is rendered at 32 timestamps from 10 cameras, giving 320 images at $256^2$.
MoVieS and 4DGT decode inside their renderers, and C4G decodes all timestamps during encoding.

\begin{table}[!htbp]
\centering
\small
\setlength{\tabcolsep}{2.5pt}
\caption{\textbf{Runtime and memory per clip.} For \ours{}, rendering includes stored-texture re-projection.}
\label{tab:runtime}
\resizebox{\columnwidth}{!}{%
\begin{tabular}{lcccccc}
\toprule
Method & Encode & Decode & Render & Per image & Memory & Params \\
\midrule
MoVieS (bf16) & 2.13\,s & -- & 2.42\,s & 7.6\,ms & 15.7\,GB & 1041M \\
4DGT & 1.10\,s & -- & 7.55\,s & 23.6\,ms & 9.8\,GB & 1223M \\
C4G & 2.29\,s & -- & 0.09\,s & 0.3\,ms & 5.3\,GB & 1014M \\
\ours{} (fp32) & 0.40\,s & 1.70\,s & 1.84\,s & 5.8\,ms & 4.8\,GB & 836M \\
\bottomrule
\end{tabular}}
\end{table}

\section{Evaluation details}

\subsection{Split and camera separation}
\label{app:split}

We compute a DINOv2 environment feature for each scene and cluster the 3{,}400 scenes with $k$-means ($k=120$).
Test scenes are drawn from clusters assigned to the test side, and each has cosine similarity below 0.62 to every scene in the remaining clusters.
This gives 101 test scenes, each rendered at four focal lengths (404 clips).

The remaining 3{,}299 scenes (13{,}196 clips) form the training set.

The four focal lengths are 18, 24, 35 and 50\,mm.
The ten cameras coincide at frame 0 and separate over time.
Measured over 404 test clips, mean pairwise separation is 0.93\,m at frame 31 and 2.10\,m at frame 80.
In first-32, 70\% of target images are within 0.5\,m of the source camera; in random windows, 35\% are.
The two protocols therefore cover different view-separation regimes.

\subsection{Visibility masks and image metrics}
\label{app:metrics}

MultiCamVideo provides no co-visibility masks.
We construct a shared mask from estimated depth and foreground segmentation, independently of each reconstruction method.
A target pixel is back-projected using its estimated depth.
Depth scale is aligned to the source using a median depth ratio before projecting the point into source frames.

A background pixel is accepted if at least one source frame sees a background surface at the projected position with depth agreement within 10\%.
An RMBG foreground pixel is accepted only through the source frame at the same time.
This restriction avoids treating a moving point as static across frames, but does not identify foreground observations connected by motion.
Sky and points beyond 40\,m are checked by direction.
The resulting masks cover on average 83\% of target pixels for first-32 and 64\% for random windows.

For mask $M$ and RGB intensities in $[0,1]$, we use
\begin{equation}
\mathrm{mPSNR}=-10\log_{10}
\frac{\sum_{\mathbf{x}}M(\mathbf{x})\|\hat I(\mathbf{x})-I(\mathbf{x})\|^2}
     {3\sum_{\mathbf{x}}M(\mathbf{x})}.
\end{equation}
mSSIM and mLPIPS average their spatial metric maps inside the mask.
Images with less than 1\% observed pixels are omitted from masked averages, about 2\% of random-window images.
Masked and uniform-fill scores use stored 8-bit renderings.

For region analysis, current-frame dynamic input voxels are rendered as camera-facing disks into each target camera.
We threshold opacity at 0.5 and dilate by two pixels.
The static region lies inside the scene cube.
Dynamic-region LPIPS uses the region's bounding box, padded by eight pixels and resized to $128^2$, rather than a masked full-image LPIPS map.

\subsection{Baseline configurations}
\label{app:baselines}

MoVieS receives all 32 frames at $224^2$, together with ground-truth cameras and timestamps.
4DGT receives all 32 frames at $560^2$ and ground-truth cameras.
Its timestamps use frame index divided by five rather than the dataset's physical rate of 15\,fps.
We select this configuration using source-view reconstruction quality only, without access to target-view ground-truth metrics.

C4G is pose-free and receives 12 frames spread uniformly over the window at $224^2$, matching its training context count.
We align its arbitrary reconstruction scale to the estimated depth of its first context frame using a median ratio.
Target cameras are expressed relative to that frame.
The fine-tuned model also keeps C4G's architecture and 12-frame input.

\subsection{DyCheck preprocessing}
\label{app:dycheck}

We use the \emph{apple}, \emph{block}, \emph{paper-windmill}, \emph{space-out}, \emph{spin}, \emph{teddy} and \emph{wheel} sequences.
The handheld video supplies 32 input frames from every second timestamp of a non-overlapping 64-timestamp window.
At 30\,fps this gives the 15\,fps input sampling used in training.
Targets are fixed-camera images present in the official validation split at those input times.
Windows with no target image are dropped.

We centre-crop the half-resolution $360\times480$ images to $360^2$ and resize to $256^2$.
Images are resampled to the intrinsics of the first handheld frame in each window.
Official co-visibility masks use nearest-neighbour resampling.
Source depth and foreground masks are prepared as for MultiCamVideo.
The baselines use Appendix~\ref{app:baselines}' settings; all receive known cameras except pose-free C4G.

DyCheck supplies LiDAR depth for the input camera, not the fixed evaluation cameras; no target-view LiDAR comparison is reported.

\begin{table}[!htbp]
\centering
\small
\caption{\textbf{mPSNR per DyCheck sequence.} The texture-free variant is an ablation of \ours{}.}
\label{tab:dycheck_seq}
\begin{tabular}{lccc}
\toprule
Sequence & MoVieS & \shortstack{w/o stored\\texture} & \ours{} \\
\midrule
apple          & \second{15.54} & \best{15.62} & 15.38 \\
block          & 14.84 & \best{15.37} & \second{15.06} \\
paper-windmill & \best{15.79} & \second{14.73} & 14.44 \\
space-out      & \second{14.16} & \best{14.36} & 14.07 \\
spin           & \best{13.49} & \second{13.48} & 13.11 \\
teddy          & 11.38 & \best{12.76} & \second{12.23} \\
wheel          & \best{12.94} & \second{11.72} & 11.46 \\
\bottomrule
\end{tabular}
\end{table}

\begin{table}[!htbp]
\centering
\small
\setlength{\tabcolsep}{3pt}
\caption{\textbf{Stored texture on real videos.}
All rows use the same model weights and inputs.
DyCheck uses official masked metrics; Neu3D uses full-image metrics over nine target cameras.}
\label{tab:texture_transfer}
\begin{tabular}{lcccc}
\toprule
& \multicolumn{2}{c}{DyCheck}
& \multicolumn{2}{c}{Neu3D} \\
\cmidrule(lr){2-3}\cmidrule(lr){4-5}
Configuration
& mPSNR$\uparrow$
& mLPIPS$\downarrow$
& PSNR$\uparrow$
& LPIPS$\downarrow$ \\
\midrule
\oursBase{}   & 14.16 & 0.617 & 16.65 & 0.394 \\
\oursStatic{} & 13.98 & 0.555 & 16.59 & 0.319 \\
\ours{}       & 13.82 & 0.543 & 16.57 & 0.291 \\
\bottomrule
\end{tabular}
\end{table}

\subsection{Neu3D preprocessing}
\label{app:neu3d}

We evaluate six Neu3D~\cite{li2022neural3dvideo} scenes using 24 non-overlapping 32-frame windows.
For each window, a fixed central camera supplies the monocular input.
At the same timestamps, the targets are cam00 and eight cameras sampled at approximately uniform index intervals, for nine target cameras in total.
We report full-image PSNR, SSIM and LPIPS over all nine targets, and PSNR separately for cam00.
Cam00 is the conventional Neu3D test camera and is approximately 10\,cm from the input camera.
All methods use the settings in Appendix~\ref{app:baselines}.

\subsection{Qualitative selection}
\label{app:qualitative}

For each protocol, clips are ranked by how close their paired PSNR difference between \ours{} and MoVieS lies to the median difference.
The figures show the two highest-ranked clips from distinct scenes whose target view at timestamp 16, taken from the camera at median distance from the source, has at least 60\% observed pixels and an observed foreground region of at least 1{,}024 pixels.
The dynamic crop is centred on the largest observed foreground component and the static crop is the most textured observed background window; both are computed from the ground truth and the visibility mask and are shared by all methods.

\section{Depth consistency}
\label{sec:geom}

MultiCamVideo does not provide ground-truth depth.
We compare rendered target-view depth with per-camera DA3 estimates using AbsRel, RMSE and $\delta_1$, the fraction of pixels whose depth ratio is within 1.25~\cite{eigen2014depth}.
We also report median-scaled AbsRel and $\delta_1$.
Each method is evaluated where it renders a surface and the reference is valid; coverage is reported alongside the errors.

\begin{table}[!htbp]
\centering
\small
\setlength{\tabcolsep}{2.3pt}
\caption{\textbf{Rendered depth vs.\ the per-camera depth estimate} at target views (test set, per-image average). ``aligned'': per-image median scaling. C4G-FT: C4G fine-tuned on MultiCamVideo.}
\label{tab:geom}
\begin{tabular}{llcccccc}
\toprule
& & & & & \multicolumn{2}{c}{aligned} & \\
\cmidrule(lr){6-7}
& Method & AbsRel$\downarrow$ & $\delta_1\uparrow$ & RMSE$\downarrow$ & AbsRel$\downarrow$ & $\delta_1\uparrow$ & cov. \\
\midrule
\multirow{5}{*}{\rotatebox{90}{First-32}}
& 4DGT    & 0.624 & 0.141 & 8.56 & 0.344 & 0.621 & \second{0.92} \\
& C4G     & 0.505 & 0.538 & 6.34 & 0.476 & 0.586 & 0.91 \\
& C4G-FT  & \second{0.267} & \second{0.592} & \second{5.16} & 0.291 & 0.598 & \best{0.98} \\
& MoVieS  & 0.561 & 0.513 & 6.21 & \second{0.250} & \second{0.717} & \second{0.92} \\
& \ours{} & \best{0.157} & \best{0.848} & \best{2.66} & \best{0.115} & \best{0.885} & 0.89 \\
\midrule
\multirow{5}{*}{\rotatebox{90}{Random}}
& 4DGT    & 0.640 & 0.155 & 8.40 & 0.370 & 0.598 & 0.81 \\
& C4G     & 0.543 & 0.522 & 6.12 & 0.450 & 0.596 & \second{0.84} \\
& C4G-FT  & \second{0.356} & \second{0.524} & \second{5.43} & 0.346 & 0.545 & \best{0.97} \\
& MoVieS  & 0.615 & 0.505 & 5.84 & \second{0.272} & \second{0.704} & 0.75 \\
& \ours{} & \best{0.259} & \best{0.784} & \best{2.92} & \best{0.170} & \best{0.831} & 0.75 \\
\bottomrule
\end{tabular}
\end{table}

Rendered depth from \ours{} is more consistent with the DA3 reference than the evaluated baselines.
This should not be interpreted as independent geometric accuracy, since DA3 also provides our input depth and training supervision; median alignment removes global scale differences but not shared DA3 bias.

\section{Additional reconstruction diagnostics}
\label{app:diagnostics}

\begin{table}[!htbp]
\centering
\small
\caption{\textbf{Static and dynamic regions}, first-32 protocol. Regions are defined in Appendix~\ref{app:metrics}.}
\label{tab:region}
\begin{tabular}{lccc}
\toprule
& \multicolumn{2}{c}{Dynamic} & Static \\
\cmidrule(lr){2-3}\cmidrule(lr){4-4}
Method & PSNR$\uparrow$ & LPIPS$\downarrow$ & PSNR$\uparrow$ \\
\midrule
4DGT        & 16.19 & 0.345 & 18.56 \\
C4G         & 14.32 & 0.434 & 15.97 \\
MoVieS      & \second{19.02} & \best{0.150} & 21.79 \\
\oursBase{} & 17.53 & 0.276 & 21.43 \\
\oursStatic{} & 18.36 & 0.201 & \second{25.41} \\
4 full frames, no dyn.\ texture & 18.47 & 0.193 & \best{25.53} \\
\ours{} & \best{19.32} & \second{0.153} & \second{25.41} \\
\bottomrule
\end{tabular}
\end{table}

\begin{table}[!htbp]
\centering
\small
\setlength{\tabcolsep}{2.3pt}
\caption{\textbf{Rendering ladder} on random windows
(8 timestamps per clip). Only the \ours{} row uses stored texture.
Rows 1--3 directly render input pixels or voxels as mean-colour disks
and are diagnostic baselines rather than upper bounds.
``dyn.'' denotes the dynamic region and ``stat.'' the static region
inside the scene cube.}
\label{tab:ladder}
\resizebox{\columnwidth}{!}{%
\begin{tabular}{lccccc}
\toprule
& \multicolumn{2}{c}{Source, dyn.}
& \multicolumn{2}{c}{Target, dyn.}
& \multicolumn{1}{c}{Target, stat.} \\
\cmidrule(lr){2-3}
\cmidrule(lr){4-5}
\cmidrule(lr){6-6}
Rendering
& PSNR$\uparrow$
& LPIPS$\downarrow$
& PSNR$\uparrow$
& LPIPS$\downarrow$
& PSNR$\uparrow$ \\
\midrule
Current-frame pixels
& 27.27 & 0.048
& 14.90 & 0.338
& 14.75 \\

Input voxels, $512^3$
& 18.12 & 0.272
& 14.67 & 0.419
& 15.39 \\

Input voxels, dyn.\ $1024^3$
& 20.15 & 0.212
& 14.81 & 0.387
& 15.39 \\

Learned state, per-frame dyn.
& 20.75 & 0.209
& 15.88 & 0.344
& 18.37 \\

Learned state, temporal slots
& 20.23 & 0.218
& 15.88 & 0.347
& 18.37 \\

\ours{} (with stored texture)
& -- & --
& 16.36 & 0.264
& 19.21 \\

MoVieS
& 21.93 & 0.089
& 16.76 & 0.257
& 18.36 \\
\bottomrule
\end{tabular}}
\end{table}

\begin{table}[!htbp]
\centering
\small
\setlength{\tabcolsep}{2.1pt}
\caption{\textbf{PSNR by frame index within the window} (random windows, mean over 9 target cameras, computed from 8-bit renders).}
\label{tab:time}
\begin{tabular}{lcccccc}
\toprule
$t$ & 0 & 4 & 8 & 16 & 24 & 31 \\
\midrule
\oursBase{} & 17.71 & 17.44 & 17.18 & 16.72 & 16.38 & 16.11 \\
\oursStatic{} & \best{19.67} & 18.18 & 17.76 & 17.10 & 16.66 & 16.33 \\
\ours{} & \best{19.67} & \second{18.49} & \second{17.92} & \second{17.17} & 16.68 & 16.30 \\
All 32 full frames & \second{19.60} & \best{18.50} & \best{17.94} & \best{17.20} & \second{16.71} & \second{16.36} \\
MoVieS           & 18.18 & 17.93 & 17.67 & \second{17.17} & \best{16.81} & \best{16.50} \\
\bottomrule
\end{tabular}
\end{table}

\section{Downstream manipulation protocol}
\label{app:control}

\paragraph{Setup.}
The autoencoder receives 4,000 adaptation steps on ManiSkill scenes before policy training.
Four $256^2$ RGB-D views, sampled on a ring around the workspace, are fused into a fixed 2.4\,m cube.
Static features pool the first three frames of each observation window.
Simulator depth and robot-link silhouettes are used; SAM~2 tracks objects causally from boxes obtained by clustering non-robot points above the table, without object-pose or object-segmentation labels.
The policy also observes the end-effector pose and gripper state.
It is trained for 60,000 steps with batch size 64 and predicts eight-step action chunks using ten Euler steps, replanning after each chunk.
The additional 2,000 demonstrations randomize motion-planning waypoints and initial joint-motion prefixes.

\paragraph{Evaluation.}
Success is measured within 200 control steps over 100 initial states and three sampling seeds per trained policy.
Li et al.~\cite{li2026structured4d} report results with fixed $512^2$ cameras and joint-position control; ours use randomized cameras and end-effector increments.
Their published success rate is therefore a reference under a different protocol.

\paragraph{Scene size.}
Our counts average 306 held-out windows: 425.9 static and 235.5 dynamic cells per timestamp, or 661.4 in total; the three-frame policy context contains 1,132.5 cells.
The same encoder and reference windows are used for both data budgets.
The approximately 5,700 cells for Li et al. come from three initial states with their fixed cameras.
These are scene-latent counts under different observations, excluding policy activations and storage side information.
Feature dimensions are reported alongside counts; fewer cells do not imply a proportional reduction in memory or computation.

\fi

\end{document}